\documentclass{article}

 \usepackage[nonatbib,position,final]{neurips_2025}

\usepackage[utf8]{inputenc} % allow utf-8 input
\usepackage[T1]{fontenc}    % use 8-bit T1 fonts
\usepackage{hyperref}       % hyperlinks
\usepackage{url}            % simple URL typesetting
\usepackage{booktabs}       % professional-quality tables
\usepackage{amsfonts}       % blackboard math symbols
\usepackage{nicefrac}       % compact symbols for 1/2, etc.
\usepackage{microtype}      % microtypography
\usepackage{xcolor}         % colors
\usepackage{graphicx}

\title{Learning from the Past: How Previous Technological Transformations Can Guide AI Development}

\author{Risto Miikkulainen$^{1,2}$, Bret Greenstein$^3$, Babak Hodjat$^2$, and Jerry Smith$^4$
 $^1$Cognizant Technology Solutions and $^2$The University of Texas at Austin    
}

\author{%
  Risto Miikkulainen\\
  The University of Texas at Austin and\\
  Cognizant AI Lab\\
  \texttt{risto@cs.utexas.edu}
  \And
  Jerry Smith\\
  Ailevate \\
  Washington, DC\\
  \texttt{jerry.smith@ailevate.com}
  \And
  Babak Hodjat\\
  Cognizant AI Lab\\
  San Francisco, CA \\
  \texttt{babak@cognizant.com}\\[-3ex]
}

\begin{document}
\maketitle
\begin{abstract}
  Artificial Intelligence (AI) is rapidly changing many areas
  of society. While this transformation has tremendous potential,
  there are several challenges as well. Using the history of computing
  and the world-wide web as a guide, in this paper we identify
  pitfalls and solutions that suggest how AI can be developed to its full
  potential. If done right, AI will be instrumental in achieving the
  goals we set for the economy, the society, and the world in general.
\end{abstract}

\section{Introduction}

In the past 70 years, Artificial Intelligence (AI) has moved from an academic
research discipline to a technology that affects people's lives every
day. For instance, humans now interact with large language models (LLMs) routinely
to get information and to solve problems; AI systems make medical 
accurate medical diagnoses; cars drive themselves in regular traffic.

At the same time, it is still unclear
how AI should be integrated into society. There are few standards,
little cooperation across companies and countries, and business users
and consumers rely on a few people who understand the technology well
enough to make judgment calls about it.  There are significant issues
that need to be solved to ensure that as AI adoption grows, it creates
positive effects on businesses and society.  Like other technologies,
AI is vulnerable to security and privacy risks, and as a learning
system, it is subject to potential biases and abuses that can cause
significant physical and financial damage.  It is now an urgent time 
to nurture AI towards maturity and responsibility.

Given that AI seems to be everywhere already, stopping to think about
how it should be harnessed right seems like a daunting task at
first. However, a crucial observation is that while AI is complex and
pervasive, it is not that different from other technologies that
started as limited research projects and ended up becoming part of
modern infrastructure, such as computing and the world-wide web (WWW).
By analyzing how these earlier technologies developed, we can gain
insight into how AI may be governed effectively in the future. By
identifying what went well and what went wrong, we may be able to
identify the dangers of AI development and adoption, and how to
mitigate those dangers in the future.

To that end, this paper first reviews the history of computing and
WWW, identifying four main stages of development in both:
Standardization, Usability, Consumerization, and Foundationalization
(Table~\ref{fg:timeline}). The same four stages are then identified
in AI, and action recommendations are made based on lessons learned in
computing and the WWW. In the end, {\bf assuming that AI is allowed to
be developed in this guided manner, it is possible to reach a better
future where AI plays a foundational role in the same way computing and
WWW do in our current society.}

\begin{table}[!t]
\centering
\begin{tabular}{|c|c|c|c|c|} 
\hline
&\textcolor{blue}{Phase 1:}&\textcolor{blue}{Phase 2:}&\textcolor{blue}{Phase 3}:&\textcolor{blue}{Phase 4}:\\
&\textcolor{blue}{Standardization}&\textcolor{blue}{Usability}&\textcolor{blue}{Consumerization}&\textcolor{blue}{Foundationalization}\\
\hline
\textcolor{blue}{Computing}&Open&GUI interfaces&iPhone and&Cloud computing\\
&Architectures&&App store&\\
\hline
\textcolor{blue}{WWW}&HTML&Stylesheets&Web 2.0 /&Digital economy\\
&&separating content&Social networks&\\
&&from presentation&&\\
\hline
\textcolor{blue}{AI}&Safety,&Trustworthiness&AI built by&Intelligence-\\
&Multiagent&&consumer&based society\\
\hline
\textcolor{red}{Best policy}&\textcolor{red}{Ensure safety and}&\textcolor{red}{Make systems}&\textcolor{red}{Ensure open}&\textcolor{red}{Establish human-}\\
\textcolor{red}{actions}&\textcolor{red}{interoperability}&\textcolor{red}{easy to use}&\textcolor{red}{innovation}&\textcolor{red}{defined objectives}\\
&&\textcolor{red}{without harm}&\textcolor{red}{before regulation}&\textcolor{red}{to ensure}\\
&&&&\textcolor{red}{positive impact}\\
\hline
\end{tabular} 
\vspace*{2ex}
\caption{\small {\bf The four phases in the development of computing and
  the WWW, and their analogies in AI.} In terms of Computing and the WWW,
  \emph{Phase 1} represents the first step           
  in commercialization.  Companies rushed to monetize years of                  
  previous research and development with initial offerings.  These              
  offerings generally did not work with offerings from other                    
  companies.  They could only be used by experts.  Also, they were              
  fragile when applied to business, requiring significant support and           
  tuning.  Companies adopted de facto and government standards to try           
  to resolve these issues and grow the value of the initial                     
  offerings. \emph{Phase 2} represents a need to break past the early           
  adopters and experts to get wider use in business.  In this phase,            
  companies focused heavily on usability of their products.  In some            
  cases, there was substantial copying of usability innovations, often          
  without any interoperability between companies.  User growth grew             
  dramatically.  \emph{Phase 3} represents the expansion of these               
  technologies to consumers and individual innovators to create value.          
  Access to app stores and self-publishing platforms dominated, and             
  user-based content grew exponentially.  \emph{Phase 4} represents a           
  turning point where businesses built natively for the new                     
  technologies begin to dominate the market.  In this phase, companies          
  that did not adapt to completely embrace the new technologies                 
  failed, while newer or more adaptive businesses capture the bulk of           
  the market value.  These same phases can be anticipated in the development
  of AI. The main difference is that usability is ahead of standardization
  in AI, creating a compelling need for regulation. Regulation must, however,
  be balanced with open innovation in the consumerization phase, so that AI 
  can develop into the empowering foundational role it can eventually serve in society.}
\label{fg:timeline}
\end{table}

\section{The Case of Computing}

Until the early 1970s, computing technology was accessible to only a
handful of individuals working in research institutions. In the mid to
late 1970s, the first personal computers were created
\cite{ceruzzi:book03,freberger:book00}. Initially they included many
different architectures and operating systems, such as Altair 8800,
Commodore PET, and Apple II, and TRS-80. However, personal computing remained a
relatively rare opportunity until the IBM PC was introduced in
1981. It had an open architecture, which made it possible for many
manufacturers and software developers to build their own machines and
systems \cite{miller:pcmag11}. This standardization resulted in PCs
becoming commonplace and useful.

PCs were still difficult to use however, until graphical user
interfaces were developed for them. Although many had existed
before, the Apple Macintosh in 1984 made such interfaces easy to use
and available to ordinary users. The Windows system soon followed for
PCs.  Through these GUIs, it was no longer necessary to be a computer
scientist to use computers. Such usability vastly expanded their
applications.

The next phase was ushered in by smartphones, and in particular the iPhone
in 2007. While early smartphones such as Nokia's Communicator and RIM
Blackberry were miniature computers, iPhone instead provided an
interface that hid the computing and focused on access to applications
\cite{molla:recode17}.  There were significant court battles around
patents (Apple vs.\ Samsung, Apple vs.\ Qualcomm), emphasizing
where innovations were creating substantial value. This phase resulted
in the consumerization of computing: Anyone could now access computing at
any time, and most of the applications became computer-oriented instead
of business-oriented.

The fourth phase is happening now: Computing has become an infrastructure
that is invisible to most people, like electricity or plumbing. It is
accessed through numerous devices, including phones, cars, cashiers,
homes, and much of it happens in the cloud instead of locally. People
do not have to care where and how it happens---they simply interact
with the results, the same way we interact with a light switch or a
faucet. Computing is a foundational infrastructure upon which we build
our everyday activities. On the other hand, it is dominated by only a
few players (AWS/MS/Google) that thereby have significant control of
those activities.

\vspace*{1.5ex}
\section{The Case of the World-Wide Web}

The development of the World-Wide Web followed a remarkably similar
trajectory. In the early 1990s, it had become possible to distribute
information over the internet, including services such as ftp sites,
Usenet news groups and gopher servers. However, they were accessible
only to the initiated and were of limited use. The first stage,
standardization, occurred with the invention of the HTML protocol
\cite{bernerslee:w393}. It made it possible to build content in a
common format, and access it worldwide using HTML readers, i.e.\
browsers.

The content was still mostly text, and therefore of limited use. The
invention of style sheets \cite{lie:book99} made the WWW vastly more
flexible and usable. Separating presentation and content, it was
possible to present information in a visual manner that made the
content accessible to the wider population. It became possible to
develop web interfaces for businesses, and information in general
became accessible through the web, in addition to traditional media.

The third step became known as Web 2.0: instead of being passive
consumers of information, anyone could now contribute to the WWW
\cite{oreilly:web2005}. People started to put much of their lives on
the WWW through social media platforms such as Facebook. It became
possible to create content such as blogs and videos, and find an
audience through YouTube and other similar sites.

Transition to the fourth phase is now almost complete. It happened
faster than expected, expedited to a
large extent by the pandemic. The WWW has become a standard
infrastructure for commerce and creativity. It is possible for small
businesses to reach consumers across the globe. Brick and mortar
stores have become secondary in many areas of retail such as books,
apparel, and groceries. Travel, entertainment, and everyday life in
general is organized through the Web. As a result, WWW is the
foundation of most human activity in the modern world, making our
lives richer and more efficient than ever before. On the other hand,
fake news now propagates easily, social computing and gaming can be
considered an addiction, and outages of social media services cause
major disruptions, demonstrating that our relationship with the WWW is complex
and still evolving.

\vspace*{1.5ex}
\section{Lessons for the Future of AI}

In light of the above two examples, let us examine where we are
with AI, how it is likely to develop in the future, and what we should
do about it.  For a long time, AI development seemed to follow a
trajectory similar to computing in the 1970s or the WWW in the early
1990s: The successes were disjoint and opaque, accessible and
understandable only to experts. However, Generative AI, i.e.\ image
generation models and LLMs, changed that virtually overnight even though
we were not ready for it. It is therefore useful to compare with the
earlier technological transformations to identify what we are missing, where we might be
headed, and how to make success more likely.

\subsection{Phase 1: Standardization}
%The next step is standardization.                                              
Like the open architecture of PCs and the HTML of the WWW, we need
open standards for AI systems. Such standards are particularly important
for the agentic systems that are now starting to emerge \cite{bengio:arxiv25,google:agent25}. AI is no
longer used just to predict what will happen, but also to make
decisions and to interact with other decision makers.

First, standards are needed to make sure that such systems are safe.
The internet is difficult to secure today partly because it was not developed
with safety in mind from the beginning \cite{ambrosin:ieeecom18}. As a result,
safety measures that have been developed later, such as multiple
authentication requirements, often make it difficult to take full advantage of
the WWW.  The lesson is that safety standards
need to be developed at the same time as the AI technology,
not as an afterthought or an add-on. In this manner, those standards
need not interfere with the value that AI provides.

For instance, an important part of safety is interpretability and 
explainability, i.e.\ having the AI explain explicitly what it is doing.
However, a strict explainability standard would rule out much of modern AI,
which is often based on statistical inference. Even though such inference
is opaque, it can still be very useful. Therefore, we should establish
standards of safety not based solely on linguistic explainability, but on guarantees
that the behavior of AI does not cause harm to people, other agents, systems, or the
environment. In practice, such a certification may mean that the AI
system knows its limits and can predict with reasonable accuracy that
they are not exceeded. Such a standard will establish safety in terms 
that do not prevent us from taking advantage of machine learning.

Second, similarly to the HTML protocol, standards are needed to build
systems of multiple agents. Such systems will take advantage of
different abilities and technologies. For instance, they may
connect a vision system to a planning system and further to a language
generation system in a self-driving car that interacts with its
passengers about the destinations, route preferences, timing
constraints, and safety. Through standards, it will be possible to
connect these agents and to swap them in and out, for instance to 
replace one language with another, or general knowledge with proprietary
knowledge, or an older version with a newer version.

Since LLMs interact with natural language, it is easy to establish
basic communication between agents. However, the standards
may need to specify how alternatives and confidence are expressed in
the interaction, and what the relationships between agents are (e.g.\
are they organized into a hierarchical command structure, or into a
panel of equal experts that negotiate; are there advisory agents, 
agents that generate solutions and others that evaluate them).
Recently, first steps have been taken towards such interoperability,
but more comprehensive standards are needed \cite{anthropic:mcp25,itu:standards25}.
Eventually such standardization will leverage the successes of
many aspects of AI and make many more applications possible.

\vspace*{1ex}
\subsection{Phase 2: Usability}
%The second phase is usability.                                                 
Like GUIs and stylesheets made computing and the WWW accessible to
non-experts, AI needs interfaces that make it possible for everyone to
use it.  In this respect AI is already ahead of those
other technologies. LLMs made it possible for anyone to interact with
AI systems without any technical training or understanding of how they
work. Usability is an intrinsic part of AI, and perhaps its greatest
success so far.

However, compared to other technologies, usability in AI emerged too
soon. The technologies were not necessarily ready for all the ways
people would use AI. Consumers, and even businesses, do not have a
good understanding of AI systems' capabilities, and especially their
limitations. Because they talk like people, it is natural to think
that they think like people--that they are actual experts. Because
AI is so easily usable, it is natural to trust it more than should
be warranted, which can cause damage.

In some sense, this problem is not new. Misinformation is common on
the WWW as well. While some of it is intentional, much of it is
not---it is simply normal human gossip, just in a new form. What makes
it particularly harmful is that it can spread almost instantaneously
around the globe, and there are still no good ways to identify and
mitigate it when it becomes damaging.

Importantly, if the AI hallucination problem can be solved, AI
could provide a solution to the WWW misinformation as well. There is
already promising work on understanding when the AI is hallucinating,
how to use trustworthy sources in retrieval-augmented generation, and
how to use AI to evaluate AI
\cite{huang:medium24,lewis:neurips20,lindsey:biology25,qiu:neurips24}.
Such techniques could automate trustworthiness evaluation of AI as
well as human output, leading to a more usable WWW as well as AI.
Interestingly, such solutions lead to questions about the nature of
truth in a world with multiple perspectives and continual change.
Addressing such questions in the development
of AI is a worthy human endeavor in itself, forcing us to give
age-old philosophical discussions concrete interpretations.

\vspace*{1.5ex}
\subsection{Phase 3: Consumerization}
%The third phase is consumerization.                                            
Like the iPhone made computing available to everyone everywhere and Web
2.0 made it possible for anyone to contribute to the WWW, in this
phase it will become possible for anyone to build AI applications to
fit their needs.  We are not there yet, and in one sense the situation is just
the opposite: only a few large research labs with industry-level
resources can develop LLMs today. However,
there is a large ecosystem of people studying those LLMs and building
applications based on them, both in academia and industry. Many people
are getting into AI: the number of students wanting to study AI, 
open jobs in AI, contributions to AI conferences,
and people using AI apps has skyrocketed in the last decade or so.

One important lesson from the past is to keep innovations open.
Consider for example the browser wars of the late 1990s
\cite{hoffmann:browserwars17}. Initially there had been rapid
development and innovation among web browsers. However in the late 1990s,
Microsoft gained a dominant position by bundling its Explorer to
Windows, and in essence cutting off competition. As a result,
innovation stopped for several years, until the antitrust case and
mobile computing got it going again.

Similarly, although here are plenty of opportunities to make
money with closed-source AI, we must ensure that there are open-source
options as well.  Much of the innovation, such as that on
explainability and safety, depends on having access to the internal
representations of the LLMs and their training data and mechanisms
\cite{lindsey:biology25,qiu:neurips24}. Further, it should not be possible for one
player to force adoption of their AI technology simply because they
dominate some other part of IT, such as cloud computing. Open
innovation is what made the current surge of AI technology
possible, through publications and open-source
\cite{tensorflow2015,pytorch19,touvron:arxiv23,vaswani:neurips17}.
We must ensure that this process continues.

Consumerization does not necessarily mean that every consumer
will be able to build their own LLM. However, it should be possible
for a large part of society---startups, tech people in business,
government, education, healthcare, and even consumers---to build custom
applications based on foundational models. That is, people can routinely
conceptualize, configure, train, and deploy such systems for their
specific needs. They may include intelligent assistants
custom-developed to manage an individual's activities, finances, and
health, design customized interiors, gardens, and
clothing, and maintain buildings, appliances, and
vehicles. The assistants should also empower them to make better
decisions in their work and in their personal life, and allow them to create digital twins
that interact with other people and their AIs. The key is that such
AI deployments are configured by individuals to match their needs.

A lesson learned from the problems with privacy and fake content in
the WWW \cite{lazer:science18,reputation17} suggests that there is a
danger that this process will run amok. However, there is also great
potential in encouraging creativity and enriching people's lives. If
Facebook and YouTube had been moderated, editorialized, and regulated
from the beginning like traditional media, it is unlikely that such
creativity would have flourished. We need to be able to watch and
learn from unbridled AI development, and avoid regulating it
until it becomes absolutely necessary. Only then will it be possible
to harness the potential of democratized AI-based innovation.

\subsection{Phase 4: Foundationalization}
%The fourth phase is foundationalization.                                       
The way computing has become invisible and the WWW has become a primary
means of interaction, so will AI become ingrained into society. It
means that AI will be routinely running business operations,
optimizing government policies, transportation, agriculture,
education, and healthcare. After a successful consumerization
phase, this change does not mean that human decision making is
replaced by machines---it means that human decision making is
empowered by machines.

Note that Agentic AI, which is currently emerging, is not limited to
prediction, but by definition prescribes what decisions need to be made to
achieve given objectives. Only humans
can define what those objectives are---we cannot delegate them to AI agents. At
the macro level, we will need to decide in what kind of society we want
to live, and derive the objectives accordingly. For instance, we
may decide to maximize productivity and growth, but at the same time
minimize cost and environmental impact, and promote equal access and
diversity. AI can then be directed to discover ways in which those
objectives can be achieved.

In past and current societies, decision making is often obscured by
special interests, historical inertia, and personal agendas, and
consequently it has been difficult to prevent conflicts and promote
opportunity despite best efforts. In contrast, AI in this fourth phase
will provide the tools to bypass such factors and build the society we
want to build. Thus, for the first time in history, we will be in
control of our own fate.

\section{Conclusion}

AI is a technology that makes a better future possible. A common
misconception is that AI is something uncontrollable that leads to
disasters or will eventually take over. However, if the lessons
from computing and the WWW are followed, it will instead be developed
by humans in the service of humans. It will eventually become powerful
enough to run much of society's infrastructure, but it will get there
safely only through the phases outlined above. Each step along the way
leads to more powerful AI that serves humanity better. Our job is to
guide its development to make this process productive and safe.

% \section*{Acknowledgments}
% Thanks to Bret Greenstein for contributions to an earlier version of this paper.

\bibliographystyle{plain}
\bibliography{main.bib}

\end{document}